\RequirePackage{fix-cm}
\documentclass[twocolumn]{svjour3}          
\smartqed  
\usepackage{graphicx}
\usepackage{latexsym}
\usepackage{subcaption}
\usepackage[misc]{ifsym}
\usepackage{hyperref}
\usepackage{amssymb}
\usepackage{amsmath}
\begin{document}

\title{Cross-modal Multi-task Learning for Graphic Recognition of Caricature Face 
}



\author{Zuheng Ming{$^1$}         \and
        Jean-Christophe Burie{$^1$} \and
        Muhammad Muzzamil Luqman{$^1$}
}


\institute{\Letter \quad Zuheng Ming \at
              \email{zuheng.ming@univ-lr.fr}           
           \and
           \Letter \quad Jean-Christophe Burie \at
              \email{jcburie@univ-lr.fr}
           \and
           \Letter \quad Muhammad Muzzamil Luqman \at
              \email{mluqma01@univ-lr.fr}\\
            \and
        {$^1$} \quad Laboratory L3i, La Rochelle University, 17402 La Rochelle, France
}

\date{Received: date / Accepted: date}

\maketitle

\begin{abstract}
Face recognition of realistic visual images has been well studied and made a significant progress in the recent decade. Unlike the realistic visual images, the face recognition of the caricatures is far from the performance of the visual images. This is largely due to the extreme non-rigid distortions of the caricatures introduced by exaggerating the facial features to strengthen the characters. The heterogeneous modalities of the caricatures and the visual images result the caricature-visual face recognition is a cross-modal problem. In this paper, we propose a method to conduct caricature-visual face recognition via multi-task learning. Rather than the conventional multi-task learning with fixed weights of tasks, this work proposes an approach to learn the weights of tasks according to the importance of tasks. The proposed multi-task learning with dynamic tasks weights enables to appropriately train the hard task and easy task instead of  being stuck in the over-training easy task as conventional methods. The experimental results demonstrate the effectiveness of the proposed dynamic multi-task learning for cross-modal caricature-visual face recognition. The performances on the datasets CaVI and WebCaricature show the superiority over the state-of-art methods. The implementation code is provided here.\footnote{\href{https://github.com/hengxyz/cari-visual-recognition-via-multitask-learning.git}{https://github.com/hengxyz/cari-visual-recognition-via-multitask-learning.git}}
\end{abstract}

\section{Introduction}
\label{intro}
\begin{figure}[t]
\begin{center}
   \includegraphics[width=0.9\linewidth]{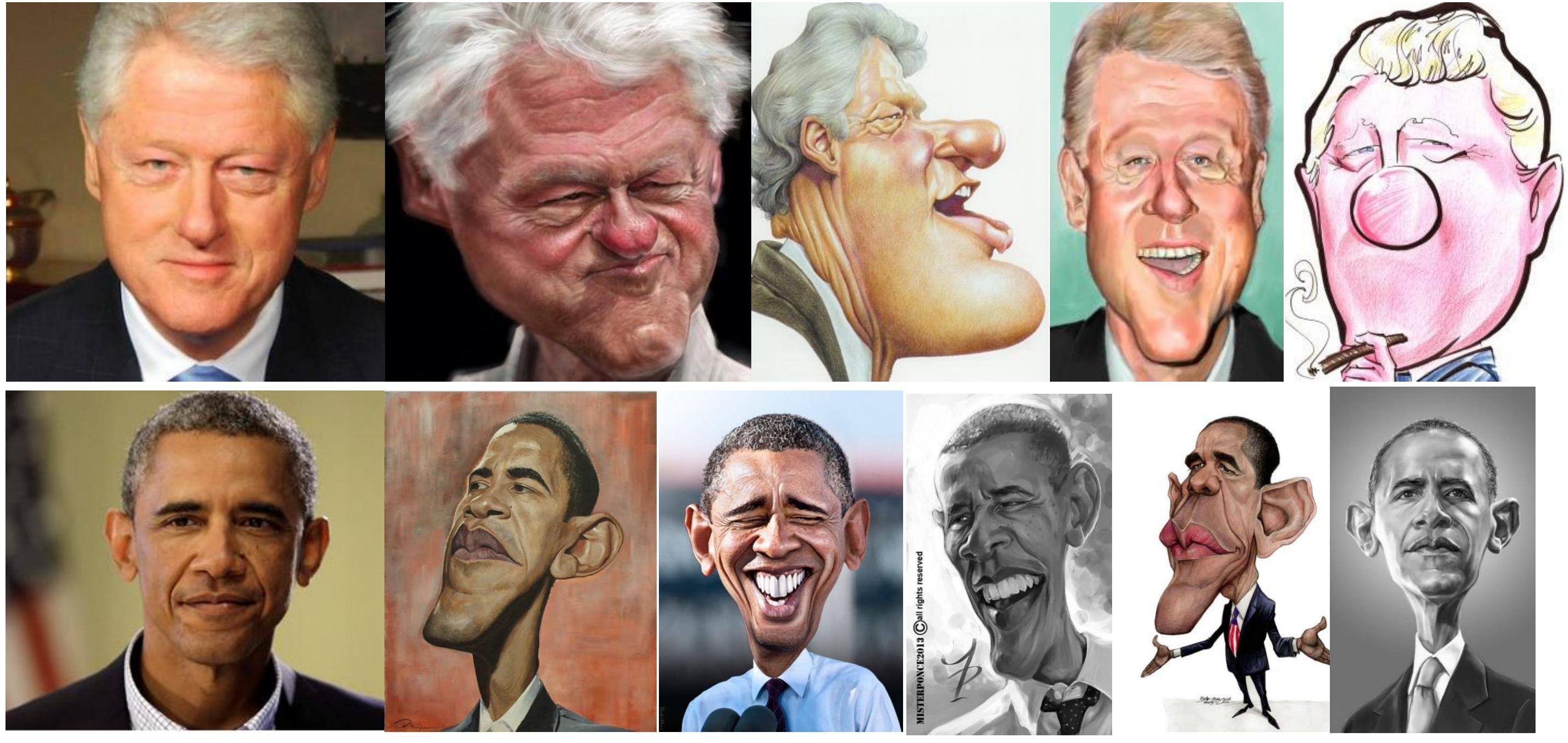}
\end{center}
   \caption{Realistic visual images and the caricatures of Bill Clinton and Barack Obama from the datasets CaVI and WebCaricature respectively. The different artistic styles result in the large variation of the caricatures even if they correspond to the same person.}
\label{fig:caricatures}
\end{figure}

\begin{figure*}
\centering
\begin{minipage}{.245\textwidth}
  \centering
  \includegraphics[width=\linewidth]{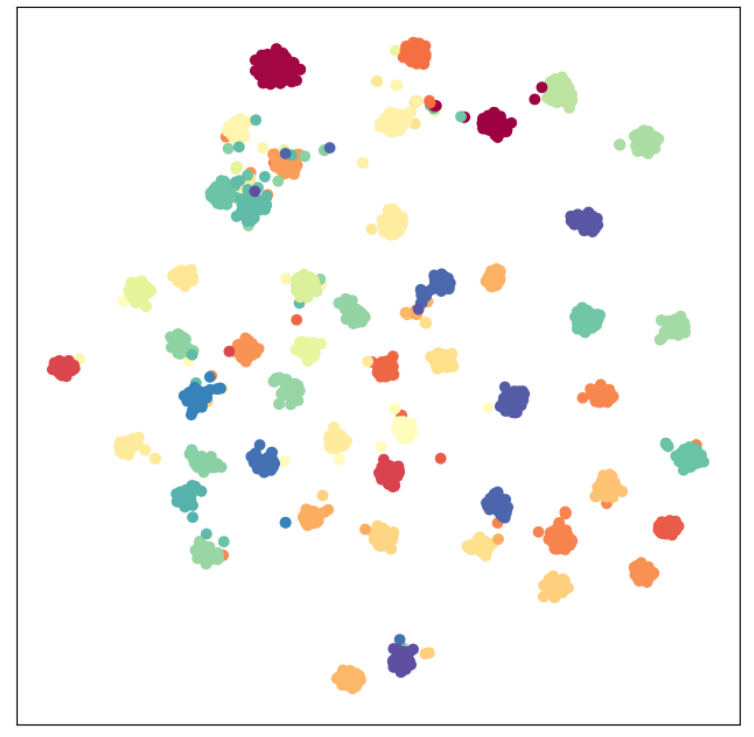}
  \subcaption{V2V classification}
  \label{fig:tsne-visual}
\end{minipage}%
\begin{minipage}{.238\textwidth}
  \centering
  \includegraphics[width=\linewidth]{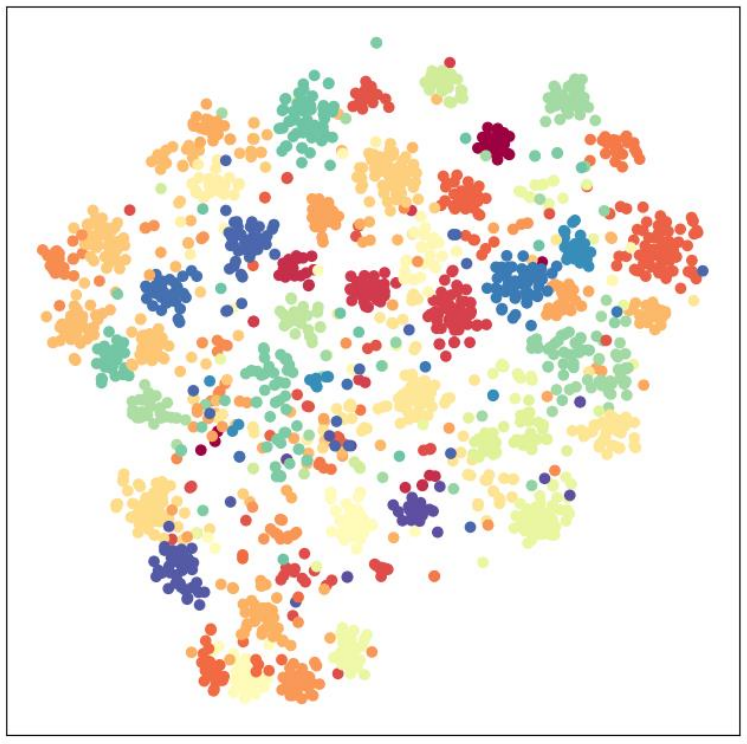}
  \subcaption{V2C classification}
  \label{fig:tsne-c2v}
\end{minipage}
\begin{minipage}{.24\textwidth}
  \centering
  \includegraphics[width=\linewidth]{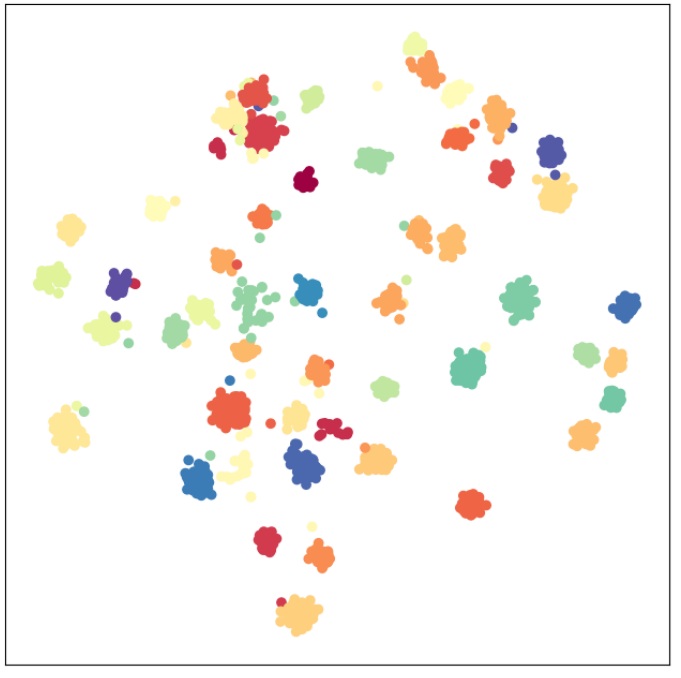}
  \subcaption{C2C classification}
  \label{fig:tsne-caricature+}
\end{minipage}
\begin{minipage}{.24\textwidth}
  \centering
  \includegraphics[width=\linewidth]{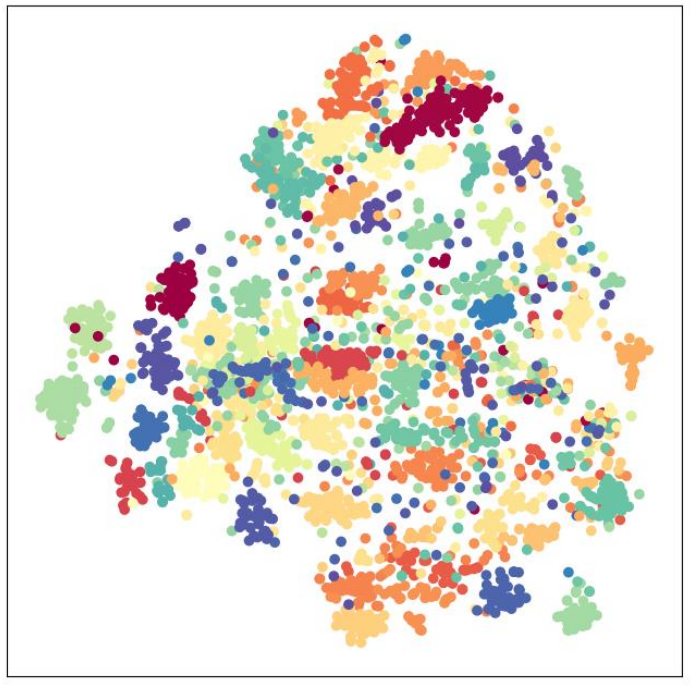}
  \subcaption{C2V classification}
  \label{fig:tsne-v2c}
\end{minipage}
\caption{Comparison of the different modalities for recognizing the realistic visual images and caricatures from the dataset CaVI \cite{garg2018deep}. The different colors denote the different 50 identities of the visual images or the caricatures. (a) V2V classification is the classification result for the visual images by the CNNs-based model trained on the realistic visual images (98.10\% of recognition accuracy); (b) V2C classification is using the same model of (a) which is trained on the visual images to classify the caricatures (53.60\% of recognition accuracy); (c) is the classification result for the caricatures by the model trained on the caricatures (78.20\% of recognition accuracy); (d) is the classification results for the real visual images by the model of (c) (41.80\% of recognition accuracy). The CNNs-based models used in all cases have the same architecture. The visualisation is implemented by t-SNE \cite{maaten2008visualizing}.}
\label{fig:t-sne-cari-visual}
\end{figure*}

In the past decade, face recognition with realistic visual image has advanced considerably. Benefiting from the powerful representation learning with the deep neural networks and particularly the deep Conventional Neural Networks (CNNs) , the performance of face recognition \cite{taigman2014deepface,parkhi2015deep,schroff2015facenet,liu2017sphereface} achieves or is beyond human being performance on datasets such as LFW \cite{huang2007labeled}, YTF \cite{wolf2011face} etc. Rather than the conventional methods based on the hand-craft feature such as LBP, Gabor-LBP, HOG, SIFT \cite{ahonen2006face,deniz2011face,bicego2006use}, the deep learning based methods mitigate the problems such as the occlusion, the illumination and the pose by leveraging the enormous data to learn the better generalized features for representing images. Nonetheless, the challenge of face recognition still exists, for instance, the non-rigid deformation and distortion as shown in the caricatures of face images. Due to the challenges of the caricature recognition, this problem is not sufficient studied.  Unlike the realistic  visual facial image, caricatures are the facial artistic drawings with the exaggerations to strengthen certain facial instinct features as shown in ~\figurename~\ref{fig:caricatures}.  With the diverse artistic styles, the caricatures are not only very different with the real visual image but also vary greatly between the caricatures with the same identity. These results both the intra-class and the inter-class variation of caricatures are quite distinct from the real visual face images \cite{HuoBMVC2018WebCaricature}. 
Thus, the visual-caricature face recognition is a cross-modal problem. It is not plausible to employ a model trained from the real visual images to recognize the caricatures and vice versa as shown in ~\figurename~\ref{fig:t-sne-cari-visual}. 
It suggests that the recognition between the realistic visual images and the caricatures is a non-trivial problem.  WebCaricature \cite{HuoBMVC2018WebCaricature} proposes a method for visual-caricature face recognition based on the pretrained VGG-Face \cite{parkhi2015deep}  with a non end-to-end framework. This method mixes the images of real visual photos and caricatures to train the final classification discriminators such as PCA \cite{kim1996face} and KCSR \cite{lei2009coupled} without considering the different recognition modalities of caricatures and visual images. The performance of this model either on face verification or on face identification is limited. Rather than the single-task method such as \cite{HuoBMVC2018WebCaricature}, the multi-task learning is particularly suitable for the caricature-visual recognition in which the different specific tasks integrated can learn the different recognition modalities as shown in \cite{garg2018deep}. 
However, the recognition on caricatures and visual images still share some common intrinsic features of face between the different modalities. This is our motivation to propose to use the hard parameter sharing structure for multi-task learning~\cite{ruder2017overview} rather than the Siamese couple networks~\cite{garg2018deep} in this work. The sharing hidden layers can share the learned common latent features between all tasks.
The multi-task learning is substantially an optimization problem for multiple objectives. While the different tasks may have different importance and also have the different training difficulty, how to find the optimal weights of tasks is an important issue in the multi-task learning. Many works prove that  the performance varies in function of the weights in the multi-task learning and the optimal performance can be obtained by the weighted task with different weights~\cite{kendall2018multi}. Thus it is unwise to assign equal weights of tasks for multi-task learning as described in~\cite{chen2017multi}. 
There are mainly two ways to search the optimal weights for multi-task learning: 1) the static method; 2) the dynamic method. In the static method, the weights of tasks are searched either manually by experimental methods such as \cite{garg2018deep,ranjan2017hyperface} or by a greedy search \cite{tian2015pedestrian}. The found optimal weights are assigned to the tasks and fixed during the training. Searching manually is laborious and ineffective while the greedy search method is time consuming. Rather than the static method, the dynamic method enables to adapt the weights automatically according to the variation of the loss,  the gradients, the uncertainty of tasks and so on \cite{kendall2018multi,zhang2016learning,chen2017gradnorm,yin2018multi}.
These methods all introduce the hyperparameters for the training of multi-task learning except \cite{yin2018multi}. However, \cite{yin2018multi} updating the dynamic weights of tasks by the total loss of the networks results in the inappropriate assignment in which the small weight is assigned to the hard task with a big loss and the large weight is assigned to the easy task with a small loss. This leads to the training of networks being stuck in the over training of the easy task and the under training of the hard task. 

\begin{figure*}[t]
\begin{center}
   \includegraphics[width=0.9\linewidth]{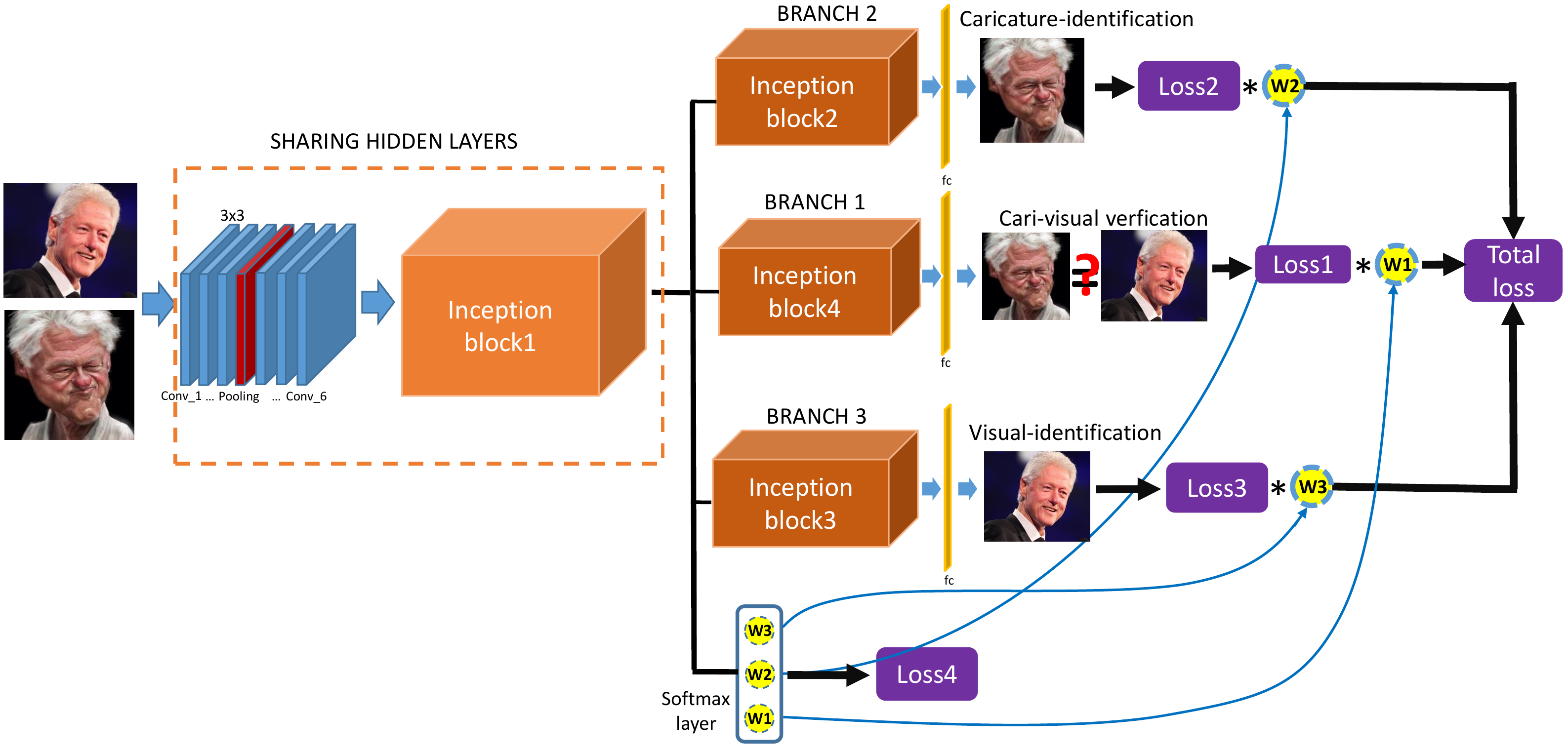}
\end{center}
   \caption{The proposed multi-task learning framework with dynamic weights of tasks for cross-modal caricature-visual face recognition. Different recognition modalities are learned by the different tasks. The introduced dynamic weights module can update the weights of tasks according to the importance of tasks during the training.}
\label{fig:framework}
\end{figure*}

In  this  work,  we  propose a  dynamic  multi-task  learning  method  based  on  the  deep CNNs to employ the caricature-visual face recognition  (see \figurename~\ref{fig:framework}). Unlike the existing approaches, the proposed method can appropriately adapts the weights of tasks according to the importance of tasks which enables the training of the networks focus on the hard task instead of being stuck in the over training of the easy task. Moreover, no hyperparameter is introduced for the training of the deep multi-task learning networks. Three different recognition tasks, i.e. caricature recognition, visual image recognition and caricature-visual face verification, with three different branches based on the sharing hidden layers are integrated in the proposed networks. Each output of this softmax layer connecting to the last layer of the hidden sharing layers serves as the dynamic weight of each task.

In summary, the main contributions of this paper are:
\begin{itemize}
	\item a multi-task learning approach with dynamic weights for the cross-modal caricature-visual face recognition, which can model the different recognition modalities by the different tasks. 
	\item a dynamic weight module without introducing additional hyperparameters can lead the multi-task learning to train the hard task primarily instead of the over training of the easy task, which results the multi-task learning more efficiently. 
    \item Both the theoretical analysis and the experimental results demonstrate the effectiveness of the proposed method for updating the dynamic weights of tasks during the training.  
    \item For all the three recognition tasks,  a multi-task learning which outperforms the state-of-the-art performances on the datasets CaVI and WebCaricature.
\end{itemize}

The remainder of this paper is organized as follows: Section II briefly reviews the related works; Section III presents the approach of multi-task learning with dynamic weights. Section IV describes the architecture of the dynamic multi-task network proposed in this work and Section V shows the experimental results. Finally, in Section VI, we draw the conclusions and present the future works.

\section{Related works}

\textbf{Caricature-Visual face recognition}  Before the representation learning by the deep CNNs, the handcrafted features such as the local descriptors HOG, LBP, Gabor-LBP, SIFT and Fisher Vector have been widely used for face recognition \cite{ahonen2006face}. By the virtue of the deep neural networks especially the deep CNNs, face recognition has made a series of breakthrough in the recent decade. DeepFace \cite{taigman2014deepface} firstly introduces a siamese network architecture for the face verification and has achieved 97.35\% on the LFW and 91.4\% on the YTF. DeepID \cite{sun2015deeply} series using more than 200 CNNs for face verification to gain a better performance (99.15\% on LFW). FaceNet \cite{schroff2015facenet} proposes triplet loss to learn embedding features for face recognition and achieve the state-of-art on LFW (99.63\%) and YTF (95.12\%). VGG face \cite{simonyan2014very} continues to implement the triplet loss on the VGG networks.  Wen and al. \cite{wen2016discriminative} propose the center loss joint with softmax to achieve the state-of-the-art performance. Recently SphereFace \cite{liu2017sphereface} proposes a revised softmax to learn "angularly" discriminative features and achieves the state-art-art performance on dataset MegaFace \cite{kemelmacher2016megaface}. Due to the challenge of the cross-modal heterogeneous face matching problem and also the lack of the dataset, the caricature-visual face recognition is not sufficiently studied especially with the deep learning based methods. Huo and al. \cite{HuoBMVC2018WebCaricature} propose a large caricature dataset called WebCaricature consisting of 252 people with 6024 caricatures and 5974 photos.  A baseline for caricature face verification and identification is also proposed respectively. It shows that the performance of the deep learning based method with pretrained VGG-Face is significantly better than the hand-craft feature based methods such as SIFT,  Gabor etc. However, the performance of the proposed method is still limited and the best performance for caricature-visual face verification is 57.22\% of validation rate (recall rate) \% @ FAR 1\%. Meanwhile, it achieves 55.41\% @ Rank-1 accuracy for caricature to real visual image identification and 55.53\% @ Rank-1 accuracy for real visual image to caricature identification. Garg et al. \cite{garg2018deep} propose the CaVINet CNN-based coupled-networks consisting of couple of 13 convolutional layers of VGGFace for cross-model caricature-verification and caricature recognition. Besides, this work also introduces a new publicly available dataset (CaVI) that contains caricatures and visual images of 205 identities, which has 5091 caricatures and 6427 visual images. The CaVINet can achieve 91.06\% accuracy for the caricature-visual face verification task, 85.09\% accuracy for caricature identification task and 94.50\% accuracy for caricature identification task. It notes that the weights of tasks are manually searched by the experimental method. 

\textbf{Multi-task learning} has been used successfully across many areas of machine learning \cite{ruder2017overview}, from natural language processing and speech recognition \cite{collobert2008unified,deng2013new} to computer vision \cite{girshick2015fast}. Fast R-CNN \cite{girshick2015fast} uses a multi-task loss to jointly train the classification and bounding-box regression for object detection. The classification task is set as the main task with the weight 1 and the bounding-box regression is set as the side task weighted by $\lambda$. The author also shows the improvement of the multi-task learning for object detection comparing to the single-task learning. Hyperface \cite{ranjan2017hyperface} proposed a multi-task learning algorithm with static weights for face detection, landmarks localization, pose estimation and gender recognition using deep CNNs. Tian et al. \cite{tian2015pedestrian} fix the weight for the main task to 1, and obtain the weights of all side tasks via a greedy search within 0 and 1. In \cite{chen2017gradnorm} the weights are updated dynamically by the loss of the gradients meanwhile an hyperparameter is introduced for balancing the training of different tasks. \cite{kendall2018multi} introduces an uncertainty coefficient $\theta$ to revise the loss function which can be fixed manually or learned based on the total loss. Zhang et al. \cite{zhang2016learning} introduce an hyperparameter $\rho$ as a scale factor to calculate the dynamic weight $\lambda_t$ of face attributes recognition.  Yin et al. \cite{yin2018multi} proposed a multi-task model for face pose-invariant recognition in which the main task is face identification and the side tasks are the classification of face pose, facial expressions and face illumination. The weight of main task is set 1 and the weights of the side tasks are assigned by the dynamic weights generated by the softmax layer. Since the dynamic weights of tasks are updated by the total loss of networks, the training of the multi-task learning  is stuck in the over training of the easy task while the hard task is under training.

\section{Dynamic multi-task learning networks}
\begin{figure*}[t]
    \begin{center}
       \includegraphics[width=0.9\linewidth]{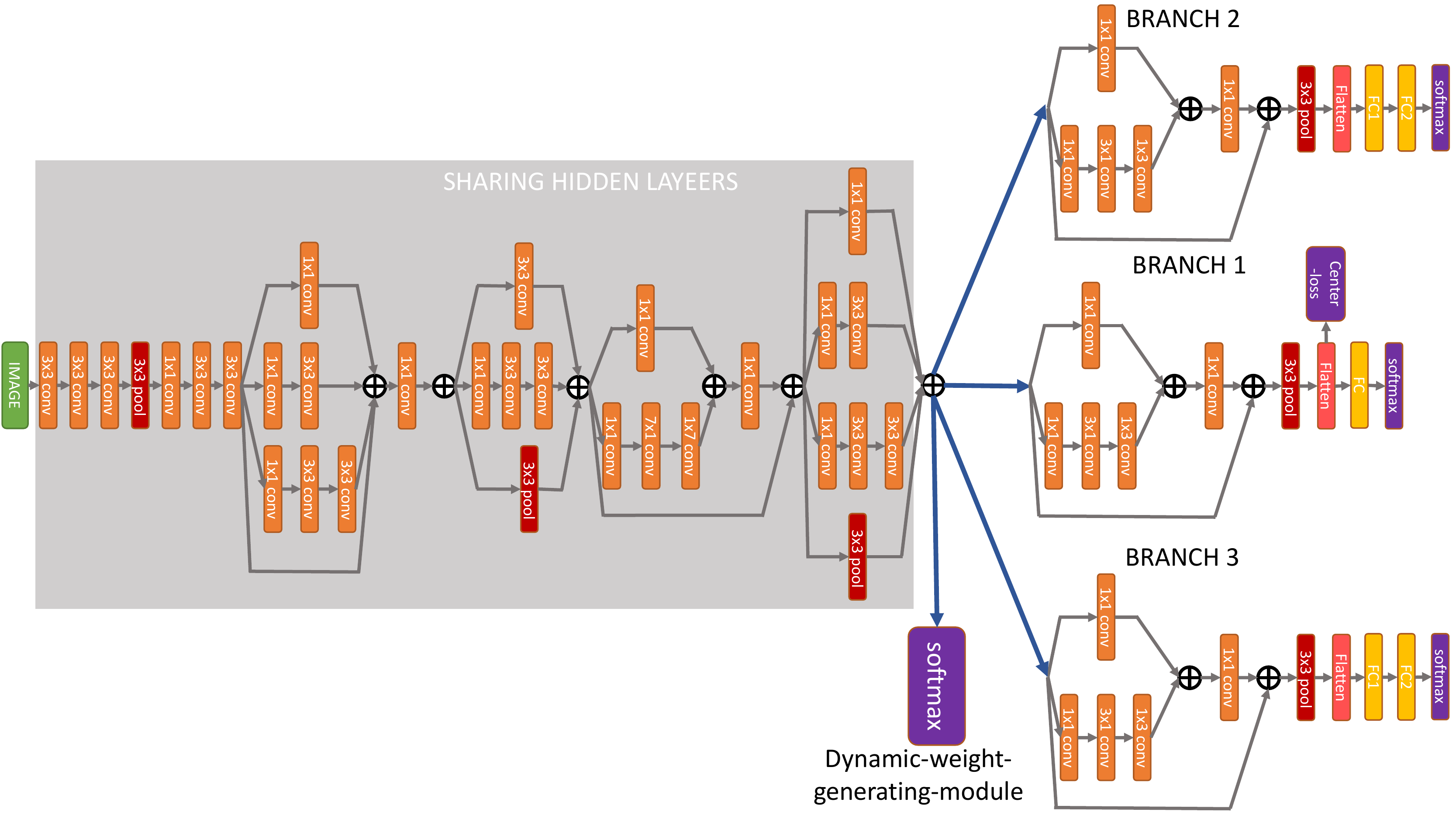}
    \end{center}
\caption{The architecture of the proposed multi-task learning networks with dynamic weights of tasks for cross-modal caricature recognition. The weights of tasks can be automatically updated by the dynamic-weights-generating-module. The networks are based on the Inception-ResNet structure.}
\label{fig:architecture}
\end{figure*}
The proposed networks for multi-task learning is based on the hard parameter sharing structure (see ~\figurename~\ref{fig:framework}), in which the sharing hidden layers can capture the modality-common features between all tasks \cite{ruder2017overview}. Although the caricature and visual images have different recognition modalities,  they still share some common features such as the face like pattern, the similar topological structure of the eyes, nose, mouth etc. Thus, we also wish the networks can learn the common features by the sharing hidden layers. 
In this work, we simplify the networks to use one-stem based networks as the hidden sharing layers to learn the common features across the different modalities, and enforce the task-specific branches to learn the modality-specific features. The deep neural networks are constructed  by the Inception-ResNet \cite{szegedy2017inception} blocks. The three branches are respectively dedicated to caricature identification, face identification and caricature-visual face verification. The three branches have almost identical structures to facilitate the transfer learning from the pretrained face recognition task. Specifically, BRANCH 1 can extract the embedded features of bottleneck layer for caricature-visual face verification, and BRANCH 2 and 3 use the fully connected softmax layer to calculate the probabilities of the identities of the input caricatures or real visual images. 
The details of the architecture of the proposed dynamic multi-task learning networks are shown in \figureautorefname~\ref{fig:architecture} and \tableautorefname~\ref{tab:architecture}. The weights of tasks are generated by the softmax layer connecting to the end of the sharing hidden layers, which can be so called the dynamic-weight-generating-module. Each element in the dynamic-weight-generating-module is corresponding to a weight of a task $w_i$, thus the size of the dynamic-weight-generating-module is equal to the number of weights of tasks, e.g. the size is 3 in this work. Since the weights are generated by the softmax layer,  $w1+w2+w3=1$ which can well indicate the relative importance of tasks. The parameters of this softmax layer are updated by the independent loss function $\mathcal{L}_4$ during the training of the networks,  which can automatically adjust the weights of tasks in light of the variation of the loss of tasks and drive the networks to always train the hard task primarily by assigning a larger weight.

\begin{table*}
\caption{\label{tab:architecture} Details of the architecture of the proposed multi-task learning networks with dynamic weights. The kernel is specified as
rows x cols x depth, stride. The repeat number of the kernel is denoted in the bracket. BRANCH denotes which branch the block belongs to. \#1x1,\#3x3,.., denote the conv kernel used in the inception block.}
\begin{center}
\small
\begin{tabular}{|l|c|c|c|c|c|c|c|c|}
\hline
Layer & Kernel & \#1x1 & \#3x3 & \#3x1 & \#1x3 & \#1x7 & \#7x1 & BRANCH\\
\hline\hline
conv1 &3x3x32,2&&&&&&&  \\\hline
conv2 &3x3x32,1&&&&&&&  \\\hline
conv3 &3x3x64,1&&&&&&&  \\\hline
maxpool1 &3x3,2&&&&&&&  \\\hline
conv4 &1x1x80,1&&&&&&&  \\\hline
conv5 &3x3x192,1&&&&&&&  \\\hline
conv6 &3x3x256,2&&&&&&&  \\\hline
Inception(1a) &&32,1&&&&&&  \\\hline
Inception(1b) &&32,1&32,1&&&&&  \\\hline
Inception(1c) &&32,1&32,1 (2)&&&&&  \\\hline
conv7 &1x1x192,1&&&&&&&  \\\hline
Inception(2a) &&384,2&&&&&&  \\\hline
Inception(2b) &&192,1&192,1; 256,1&&&&&  \\\hline
Inception(2c) &maxpool 3x3,2&&&&&&&  \\\hline
Inception(3a) &&128,1&&&&&&  \\\hline
Inception(3b) &&128,1&&&&128,1&128,1&  \\\hline
Inception(4a) &&256,1&&&&&&  \\\hline
Inception(4b) &&256,1&256,2&&&&&  \\\hline
Inception(4c) &&256,1&256,1; 256,2&&&&&  \\\hline
Inception(4d) &maxpool 3x3,2&&&&&&&  \\\hline
Inception(5a) &&192,1&&&&&&1  \\\hline
Inception(5b) &&192,1&&192,1&192,1&&&1  \\\hline
conv8 &1x1x192,1&&&&&&&1  \\\hline
avgpool1 &&&&&&&&1  \\\hline
fullyconn1 &&&&&&&&1  \\\hline
Inception(7a) &&192,1&&&&&&2  \\\hline
Inception(7b) &&192,1&&192,1&192,1&&&2  \\\hline
conv9 &1x1x192,1&&&&&&&2  \\\hline
avgpool2 &&&&&&&&2  \\\hline
fullyconn3 &&&&&&&&2  \\\hline
fullyconn4 &&&&&&&&2  \\\hline
Inception(9a) &&192,1&&&&&&3  \\\hline
Inception(9b) &&192,1&&192,1&192,1&&&3  \\\hline
conv10 &1x1x192,1&&&&&&&3  \\\hline
avgpool3 &&&&&&&&3  \\\hline
fullyconn4 &&&&&&&&3  \\\hline
fullyconn5 &&&&&&&&3  \\\hline

\end{tabular}
\end{center}

\end{table*}

\section{Multi-task learning with dynamic weights}
\begin{figure}[t]
    \begin{center}
       \includegraphics[width=0.9\linewidth]{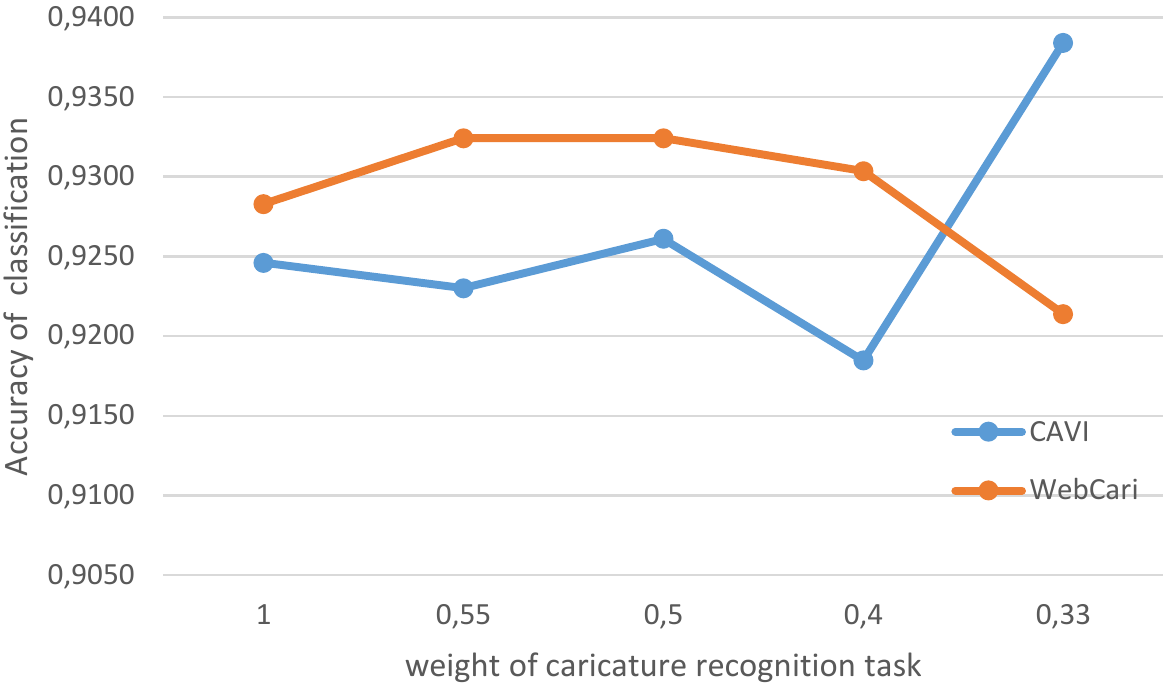}
    \end{center}
\caption{The performance (accuracy) of caricature recognition task with different weights in the multi-task learning framework. When weight of task equals 1, the networks is in single-task learning. The blue curve is the result performing on the CaVI dataset and the orange curve is the result on the WebCaricature dataset.}
\label{fig:weights-setting}
\end{figure}

The multi-task learning is an optimization  problem  for  multiple  objectives. The dominant approach used by prior works \cite{ranjan2017hyperface,tian2015pedestrian,girshick2015fast} of deep multi-task learning is to combine the different losses of the tasks with the static weights:  
\begin{equation}
\label{eq_classicMTL}
\mathcal{L}(\mathbf{X};\Theta) = \sum_{i=1}^{T} w_i\mathcal{L}_i(\mathbf{X}_i;\Theta_i)
\end{equation}
where $T$ is the number of tasks. ${X_i}$ and ${\Theta_i}$ are the input and the parameters of model. $w_i$ is the weight of tasks. Comparing to the separate networks trained on each task individually, the multi-task learning can perform better for each task by joint learning the tasks with optimal weights.  As illustrated in~\ref{fig:weights-setting}, we observe that the caricature recognition achieves the best performance when its weight is around 0.5 on dataset WebCari and around 0.3 on dataset CAVI. When the weight equals 1, the networks is in single-task learning in which the networks only trained on the caricature recognition task. We can observe that the performance of single-task learning is worse than the multi-task learning. However, Finding optimal weights is expensive, 
time-consuming and increasingly difficult for large models with numerous tasks. Here, we propose the deep multi-task CNNs with dynamic weights of tasks which are generated automatically according to the importance of tasks.

(I) \noindent~\textbf{Dynamic multi-task loss $\mathcal{L}$}: The multi-task total loss $\mathcal{L}$ is defined as follows:
\begin{equation}
\label{eq1}
\mathcal{L}(\mathbf{X};\Theta;\Psi) = \sum_{i=1}^{T} w_i(\Psi)\mathcal{L}_i(\mathbf{X}_i;\Theta_i)
\end{equation}
where $T$ is the number of tasks, here $T=3$. ${X_i}$ and ${\Theta_i}$ are the features and the parameters corresponding to each task $i$, $\Theta=\{\Theta_i\}_{i=1}^{T}$ are the overall parameters of the networks to be optimized by the total loss $\mathcal{L}$.   The parameters of the softmax layer in the dynamic-weight-generating-module is denoted as $\Psi$ which is used to generate the dynamic weights  $w_i \in[0,1]$ s.t. $\sum w_i=1$. Note that the $\Psi \not\in \Theta$.  Thus $\{\mathbf{X}_i, \Theta_i\}\in \mathbb{R}^{d_i}$, where $d_i$ is the dimension of the features $X_i$, and $\{\mathcal{L}_i, w_i\}\in \mathbb{R}^1$.  Particularly, when $w_i$ = 1 and $w_{j\neq i}$ = 0 the multi-task networks are degraded as the single-task networks. For  example, $w_1$ = 1 and $w_2$=0, $w_3$=0, is degraded to the single task network for caricature recognition (i.e. consisting of BRANCH 1 and the sharing hidden layers). 

(II)\noindent~\textbf{Caricature-Visual face verification task loss} $\mathcal{L}_1$: The loss for caricature-visual face verification task is measured by the center loss \cite{wen2016discriminative} joint with the cross-entropy loss of softmax of BRANCH 1. The loss function $\mathcal{L}_1$ is given by:      
\begin{equation}
\label{L1}
\mathcal{L}_1(\mathbf{X}_1;\Theta_1) = \mathcal{L}_{s1}(\mathbf{X}_1;\Theta_1) + \alpha\mathcal{L}_c(\mathbf{X}_1;\Theta_1)
\end{equation}

where $\mathcal{L}_{s1}$ is the cross-entropy loss of softmax of BRANCH 1, $\mathcal{L}_{c}$ is the center loss weighted by the hyperparameter $\alpha$. The $\mathcal{L}_{c}$ can be treated as a regularization item of softmax loss $\mathcal{L}_{s1}$ which is given by:
\begin{equation}
\label{eq:cross_entropy} 
\mathcal{L}_{s1}(\mathbf{X}_1;\Theta_1)= \sum^{K}_{k=1}-y_klogP(y_k=1|\mathbf{X}_1,\theta_k) 
\end{equation}

where $K$ is the number of  identities in the training dataset, ${y_k}\in\{0,1\}$ is the label of the feature $\mathbf{X}_1$, $P(y_k|\mathbf{X}_1,\theta_k)$ is softmax function. The bottleneck layer of BRANCH 1 is extracted as the feature $\mathbf{X}_1$ of the input image. The center loss $\mathcal{L}_c$ is given by:

\begin{equation}
\label{eq:centerloss} 
\mathcal{L}_c(\mathbf{X}_1;\Theta_1) = ||\mathbf{X}_1-C_{y_k}||
\end{equation}
Where the $C_{y_k}$ is the center of the class which $\mathbf{X}_1 $ belonging to, $C_{y_k} \in \mathbb{R}^{d_1}$.

(III)\noindent~\textbf{Caricature identification task loss $\mathcal{L}_2$, and Visual identification task loss $\mathcal{L}_3$} : The loss function $\mathcal{L}_2$ and $\mathcal{L}_3$ are the cross-entropy loss of the softmax layer of BRANCH 2 and BRANCH 3 respectively. The equations of $\mathcal{L}_2$, $\mathcal{L}_3$ are as same as Equation~\ref{eq:cross_entropy}, and the $K$ in $\mathcal{L}_{2}$ or $\mathcal{L}_{3}$ is the number of the identities, $\mathbf{X}_2$ or $\mathbf{X}_3$ is the bottleneck layer of BRANCH 2 or BRANCH 3. 


(IV)\noindent~\textbf{Generation of the dynamic weights $w_i(\Psi)$}:
The dynamic weights $w_i$ are generated by the softmax layer of the dynamic-weight-generating-module which is given by: 
\begin{equation}
\label{eq1_dynamicweights}
w_i(\mathbf{Z};\Psi) = \frac{e^{f^{\psi_i}(\mathbf{Z})}}{\sum^{T}_{i'}e^{f^{\psi_{i'}}(\mathbf{Z})}}
\end{equation}
where the $\mathbf{Z} \in \mathbb{R}^{d_z}$ is the flattened output of the last layer of the sharing hidden layers. $T$ is the number of tasks, here $T$=3. $\psi_i$ is parameters in the softmax layer of the dynamic-weight-generating-module $\{\psi_i\}_{i=1}^{T}=\Psi$, $\psi_i\in \mathbb{R}^{d_z}$. $f^{\psi_{i}}(\mathbf{Z})$ is activation function which is given by:https://www.overleaf.com/3624139718dhvyqdcqkgrm
\begin{equation}
\label{activefunction}
f^{\psi_i}(\mathbf{Z}) = \psi_i\mathbf{Z}^T + b_i
\end{equation}
Note that, we do not use the Relu function as the activation function since Relu discards the values minors zero. This shrinks the range of the variation of the dynamic weights $w_i$.
 
(V) \noindent~\textbf{Update of the dynamic weights $w_i$}:
We propose a new loss function to update the dynamic weights which can drive the networks always train the hard task:
\begin{equation}
\label{eq:L3} 
\mathcal{L}_{4}(\mathbf{Z};\Psi)=\sum^{T}_{i=1}\frac{w_i(\psi_i)}{\mathcal{L}_i(\Theta_i)} \quad s.t. \quad \sum w_i=1
\end{equation}
Note that, $\mathcal{L}_i\{\Theta_i\}$ is independent with $w_i(\psi_i)$ since $\Theta_i\cap \psi_i=\varnothing$ , $i \in [1,..,T]$, thus $\mathcal{L}_i$ is constant for the dynamic weight update loss function $\mathcal{L}_4$.
 
(VI)\noindent~\textbf{Analysis of the dynamic weights} Here we show how the proposed dynamic weights drive the networks focus on training the hard task. Considering the Equation~\ref{eq1_dynamicweights} and Equation~\ref{activefunction}, the gradient of the $\psi_i$ can be given by
\begin{equation}
\label{eq:gradientofweights} 
\nabla{\psi_i}=\frac{\partial \mathcal{L}_4}{\partial \psi_i} =\frac{1}{\mathcal{L}_i}\frac{\partial w_i(\psi_i)}{\partial \psi_i}= \frac{1}{\mathcal{L}_i}\frac{a_i\sum^{T}_{j\neq i}a_{j}}{(\sum_i^T a_i)^2}\mathbf{Z}
\end{equation}
where $a_i = e^{{\psi_i}\mathbf{Z}^T+b_i}$, and the update of the parameters is  $\psi_i^{t+1}=\psi_i^{t}-\eta\nabla{\psi_i}^t$ where $\eta$ is the learning rate. Then the new value of the dynamic weight $w_i^{t+1}$ can be obtained by the Equation~\ref{eq1_dynamicweights} with the  $\psi_i^{t+1}$. 
We assume the $b_i^0=0, \psi_i^0=0, \eta=1$,  (this is possible if we initialize the $\psi_i, b_i$ by zero), the $\psi_i^t$ can be given by
\begin{equation}
\label{eq:parametersupdate} 
\psi_i^t=-\sum\frac{1}{\mathcal{L}_i}\frac{a_i\sum^{T}_{j\neq i}a_{j}}{(\sum_i^T a_i)^2}\mathbf{Z}
\end{equation}
if we consider the case for two tasks $w_1$ and $w_2$:
\begin{equation}
\label{eq:dynamicweightsanaylsis} 
\begin{split}
\frac{w_1^t}{w_2^t}&=e^{(\psi_1^t-\psi_2^t)\mathbf{Z}^T} \\
&=e^{(\frac{1}{\mathcal{L}_2}-\frac{1}{\mathcal{L}_1})\frac{a_1a_2}{(a_1+a_2)^2}\mathbf{Z}\mathbf{Z}^T}
\end{split}
\end{equation}
We can see that $a_i>0$ and $ZZ^T\ge 0$, so if $\mathcal{L}_2 < \mathcal{L}_1$ the $\frac{w_1}{w_2}>1$ namely $w_1>w_2$. It means if the loss of task1 is larger than the loss of task 2, the weight of the task1 is larger than the one of task2. It indicates that the proposed loss function $\mathcal{L}_3$ can well update the weights of tasks to drive the networks and always train the hard task first.

(VII)\noindent~\textbf{Training protocol}:  The training of the entire deep CNNs includes two independent training: the training of the parameters of the networks {$\Theta$} by the multi-task loss $\mathcal{L}(\Theta)=\sum_{i=1}^{3}\mathcal{L}_i(\theta_i)$ and the training of the parameters of weight-generate-module {$\Psi$} by the loss $\mathcal{L}_4(\Psi)$. These  can be conducted simultaneously in a parallel way.
\begin{equation}
\Theta^{t-1}-\eta\frac{\partial\mathcal{L}(\Theta)}{\partial\Theta}\mapsto\Theta^{t}
\end{equation}
\begin{equation}
\Psi^{t-1}-\eta\frac{\partial\mathcal{L}_4(\Psi)}{\partial\Psi}\mapsto \Psi^{t}
\end{equation}
where $\eta\in(0,1)$ is the learning rate.

\section{Experiments and analysis}
\subsection{Datasets}
CaVI and WebCaricature are so far the largest public datasets for caricature-visual recognition research. In this work, the two datasets are all used to train and evaluate our proposed model.

~\textbf{CaVI} contains caricatures and visual images of 205 identities, which has 5091 caricatures ranging from 10-15 images per identity and 6427 visual images ranging from 10-15 images per identity. OpenFace \cite{amos2016openface} is used to extract faces from the scrapped visual images and verify the estimated bounding box ma\-nually to ensure the accuracy of the detected faces. The faces are ma\-nually extracted for caricatures and only the complete faces were annotated in the dataset.  

~\textbf{WebCaricature} is a large caricature dataset of 252 people with 6024 caricatures and 5974 photos is proposed. For each person, the number of caricatures ranges from 1 to 114 and the number of photos from 7 to 59. The caricatures are labeled manually with 17 landmarks and the landmarks of photos are detected automatically by the software Face++ \cite{Face++}. 
\subsection{Pretrained Model}
Both datasets CaVI and WebCaricature are relatively small to train such a deep CNNs for face recognition. Before training  the proposed multi-task CNNs, a single-task network consists of the sharing hidden layers. The BRANCH 3 is pretrained for face verification task with the large-scale dataset MSCeleb-1M \cite{guo2016ms}. MTCNN \cite{zhang2016joint} is used to detect the face from the raw images. The RMSprop with the mini-batches of 90 samples is applied for optimizing the parameters. The momentum coefficient is set to 0.99. The learning rate starts from 0.1, and is divided by 10 at the 60K, 80K iterations respectively. The dropout probability is set to 0.5 and the weight decay is 5e-5. The networks are initialized by Xavier \cite{glorot2010understanding} with the zero bias. Then, the training of the dynamic multi-task CNNs can be conducted with the initialization of BRANCH2 and BRANCH1 by the pre-trained BRANCH3.

\subsection{Toy example}
\begin{figure}
\centering
\begin{minipage}{.24\textwidth}
  \centering
  \includegraphics[width=\linewidth]{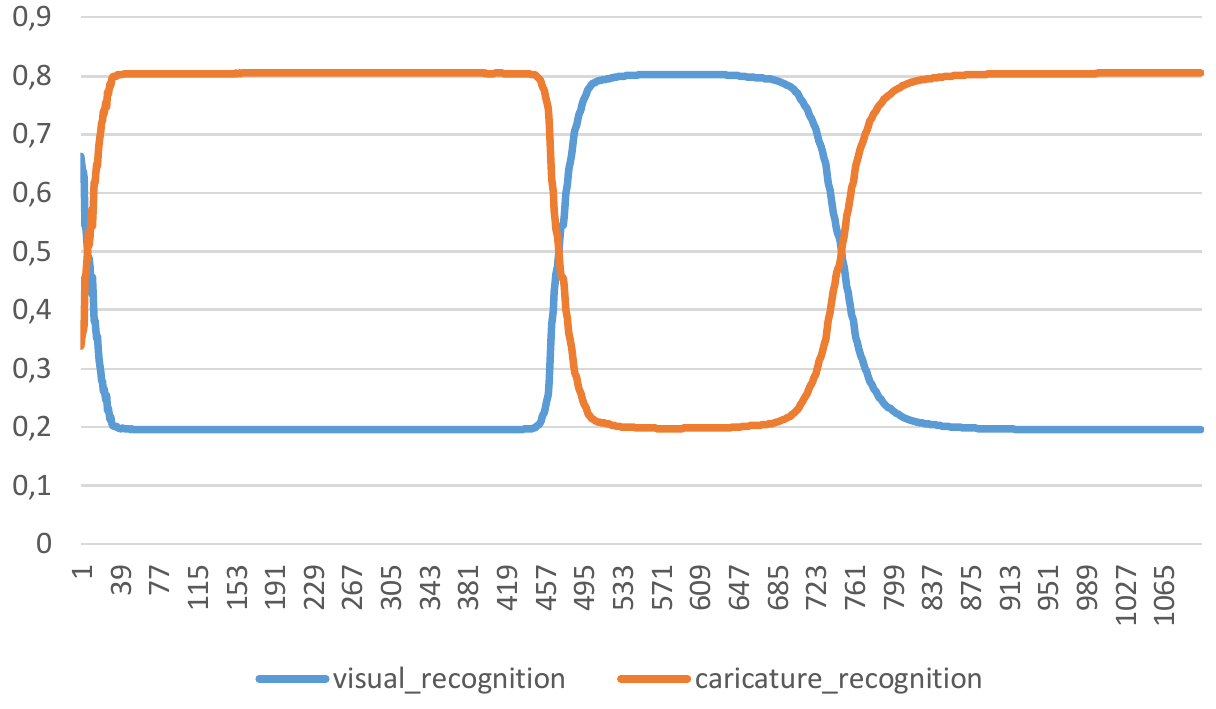}
  \subcaption{Our dynamic weights}
  \label{fig:dynweights_real_CK+_twotasks}
\end{minipage}%
\begin{minipage}{.24\textwidth}
  \centering
  \includegraphics[width=\linewidth]{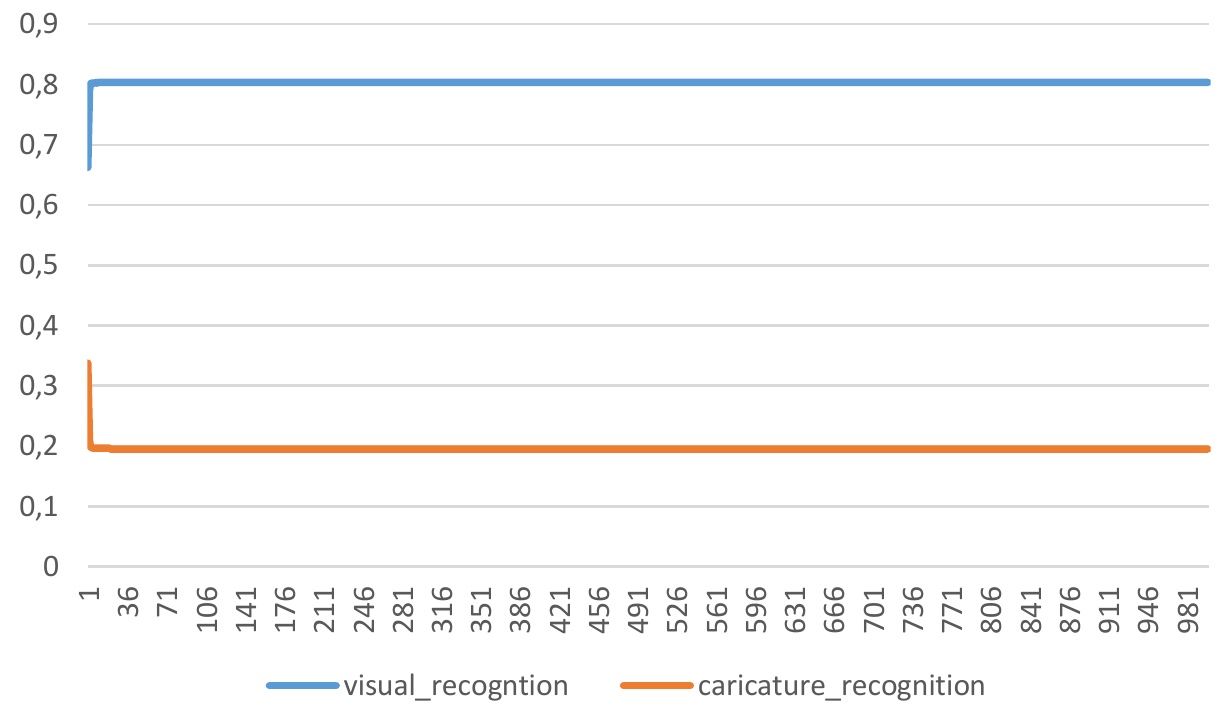}
  \subcaption{Naive dynamic weights}
  \label{fig:dynweights_real_oulu_twotasks}
\end{minipage}
\begin{minipage}{.23\textwidth}
  \centering
  \includegraphics[width=\linewidth]{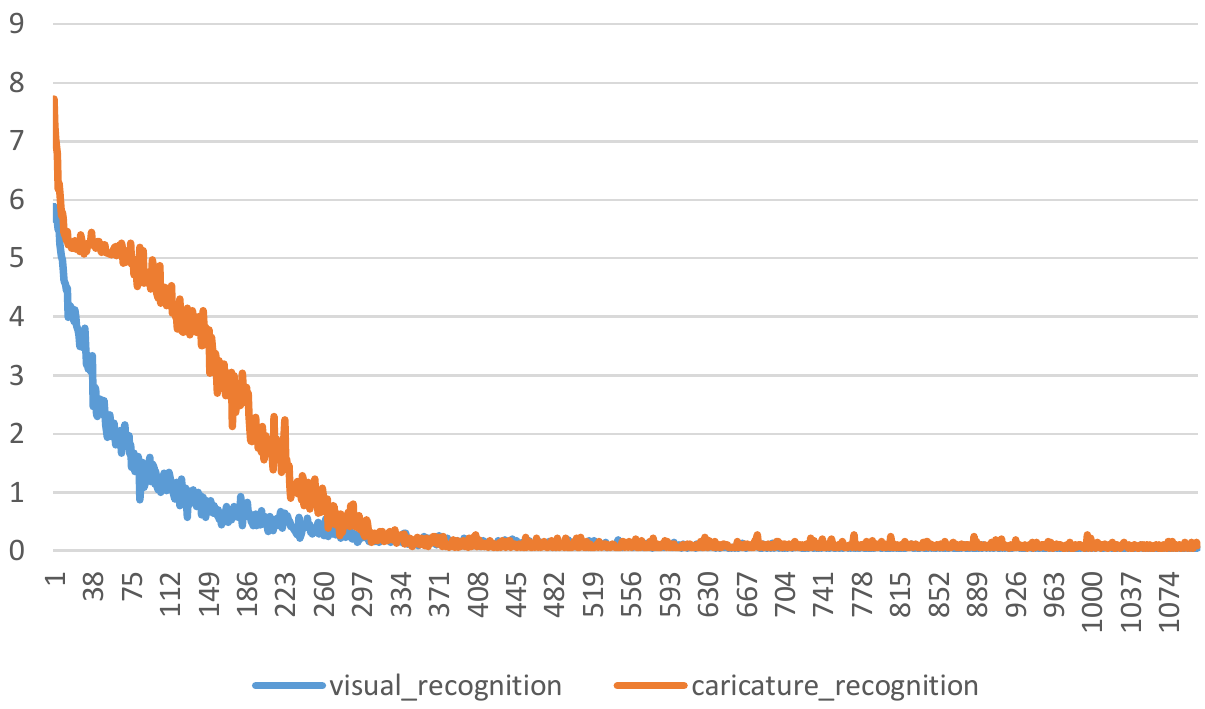}
  \subcaption{Loss of our model}
  \label{fig:dynweights_fake_CK+_twotasks}
\end{minipage}
\begin{minipage}{.23\textwidth}
  \centering
  \includegraphics[width=\linewidth]{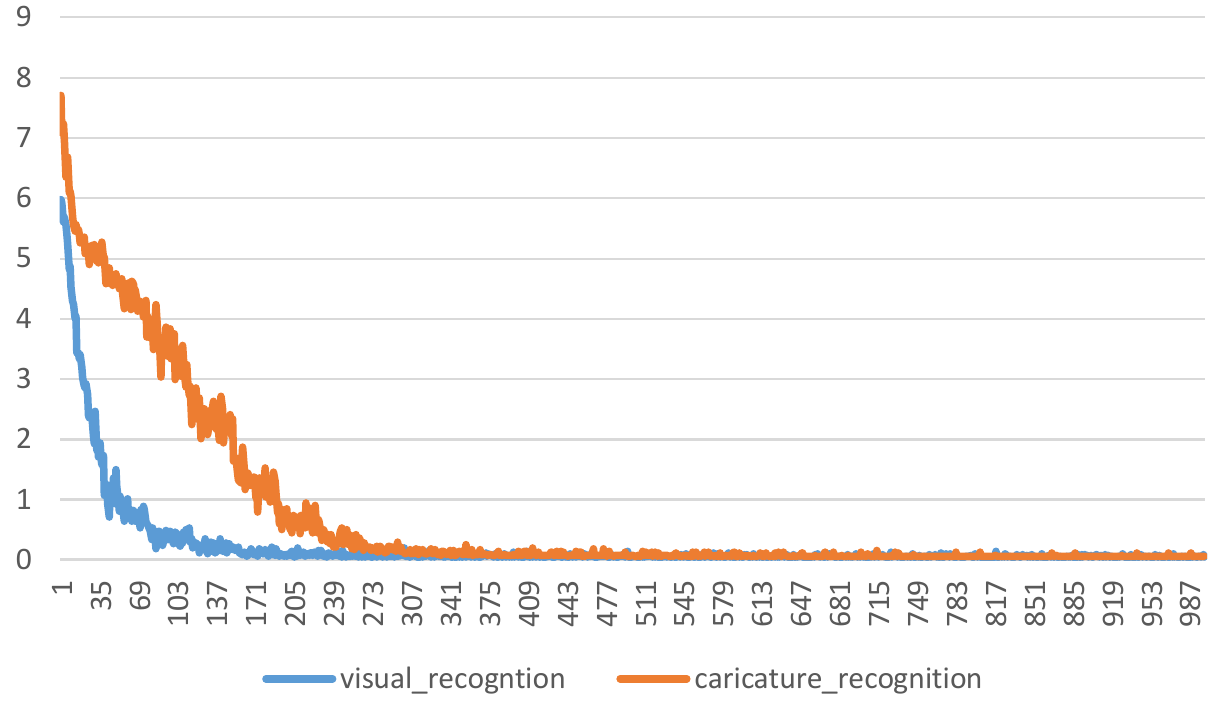}
  \subcaption{Loss of naive dynamic model}
  \label{fig:dynweights_fake_oulu_twotasks}
\end{minipage}
\caption{Comparison of our dynamic multi-task learning and the naive dynamic multi-task learning proposed in \cite{yin2018multi} for the two tasks of caricature recognition and visual image recognition on dataset CAVI. The upper row shows the dynamic weights for both methods and the bottom row is corresponding to the losses. (a), (c) are corresponding to the proposed approach and (b), (d) are the naive dynamic multi-task learning. The orange curves correspond to the caricature recognition and the blue curves denote the visual image recognition. The horizontal axis is the number of training iterations.}
\label{fig:toy-dynamicl}
\end{figure}

In order to better demonstrate the effectiveness of the proposed dynamic multi-task learning for the different recognition modalities, a toy example on two-task based multi-task learning is conducted on dataset CAVI. The caricature recognition and the visual image recognition are selected as two entirely independent tasks. Thus, the weight $w_1$ of the remaining task for caricature-visual face verification is set to 0 and $w_2+w_3=1$. In~\figurename~\ref{fig:toy-dynamicl}, we compare our dynamic multi-task learning approach and the method proposed in \cite{yin2018multi}. For convenience, the method of \cite{yin2018multi} is called naive dynamic multi-task learning. We can see that for both methods, the caricature recognition (denoted in orange) with relatively large loss is the hard task at the beginning of the training. While our  method assigns a large weight (the orange in (a)) enabling to fully train the hard task of caricature recognition. Instead, the naive dynamic method assigns a large weight to the easy visual recognition task with small loss (denoted in blue). In the following training, the dynamic weights of our method can adapt automatically according to the loss of tasks. However, the naive dynamic method is stuck in the training of the easy task since a larger weight will be always assigned to the easy task and  the lost will get smaller. 
This state can be hardly turned over unless the loss of the hard task decrease quicker than the easy task even with a very small weight. This is the reason why the naive dynamic multi-task learning results in the over training of the easy task and the under training of the hard task.


\begin{figure*}
\centering
\begin{minipage}{.238\textwidth}
  \centering
  \includegraphics[width=\linewidth]{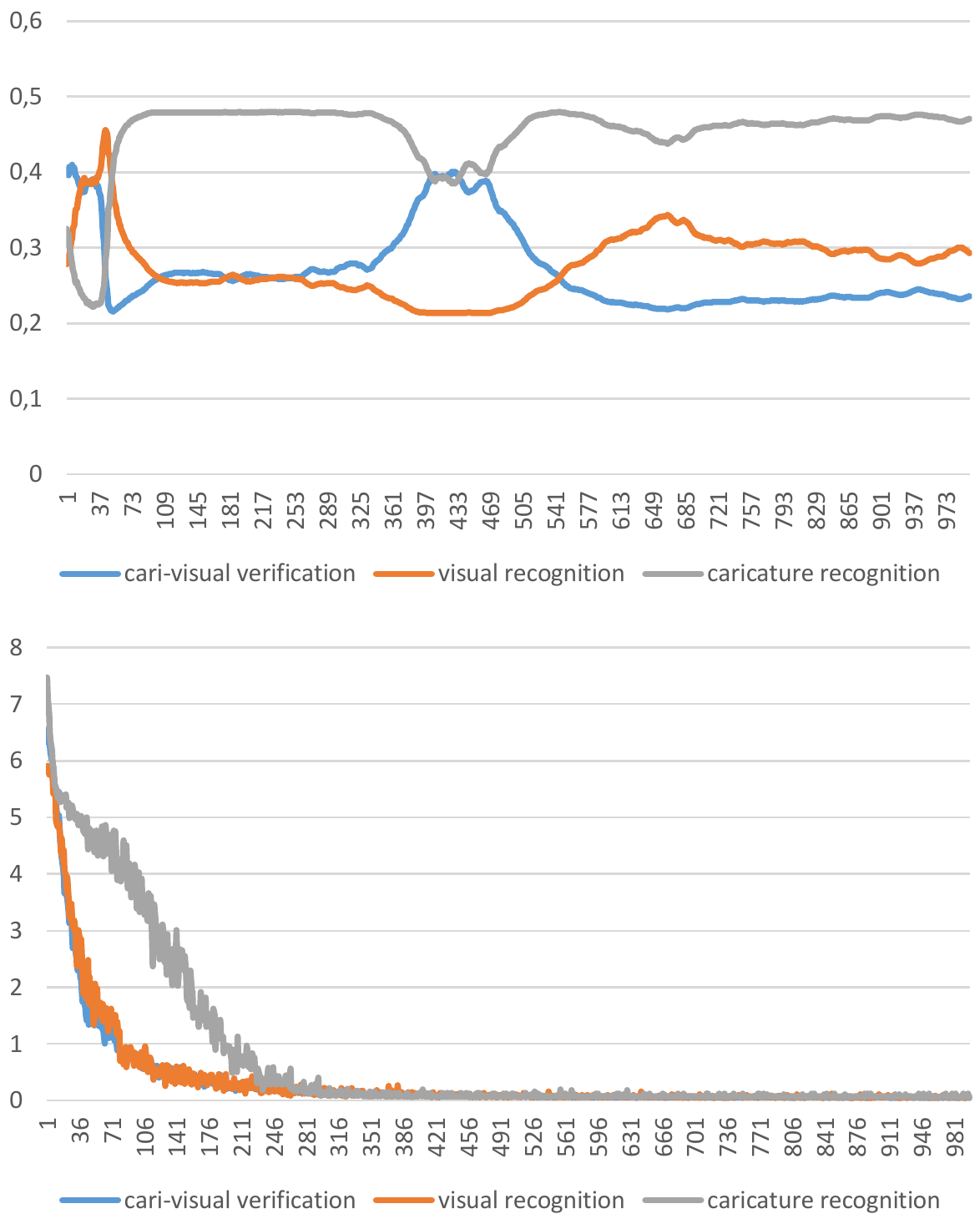}
  \subcaption{Our Dynamic-CAVI}
  \label{fig:dynweights_real_CK+}
\end{minipage}%
\begin{minipage}{.245\textwidth}
  \centering
  \includegraphics[width=\linewidth]{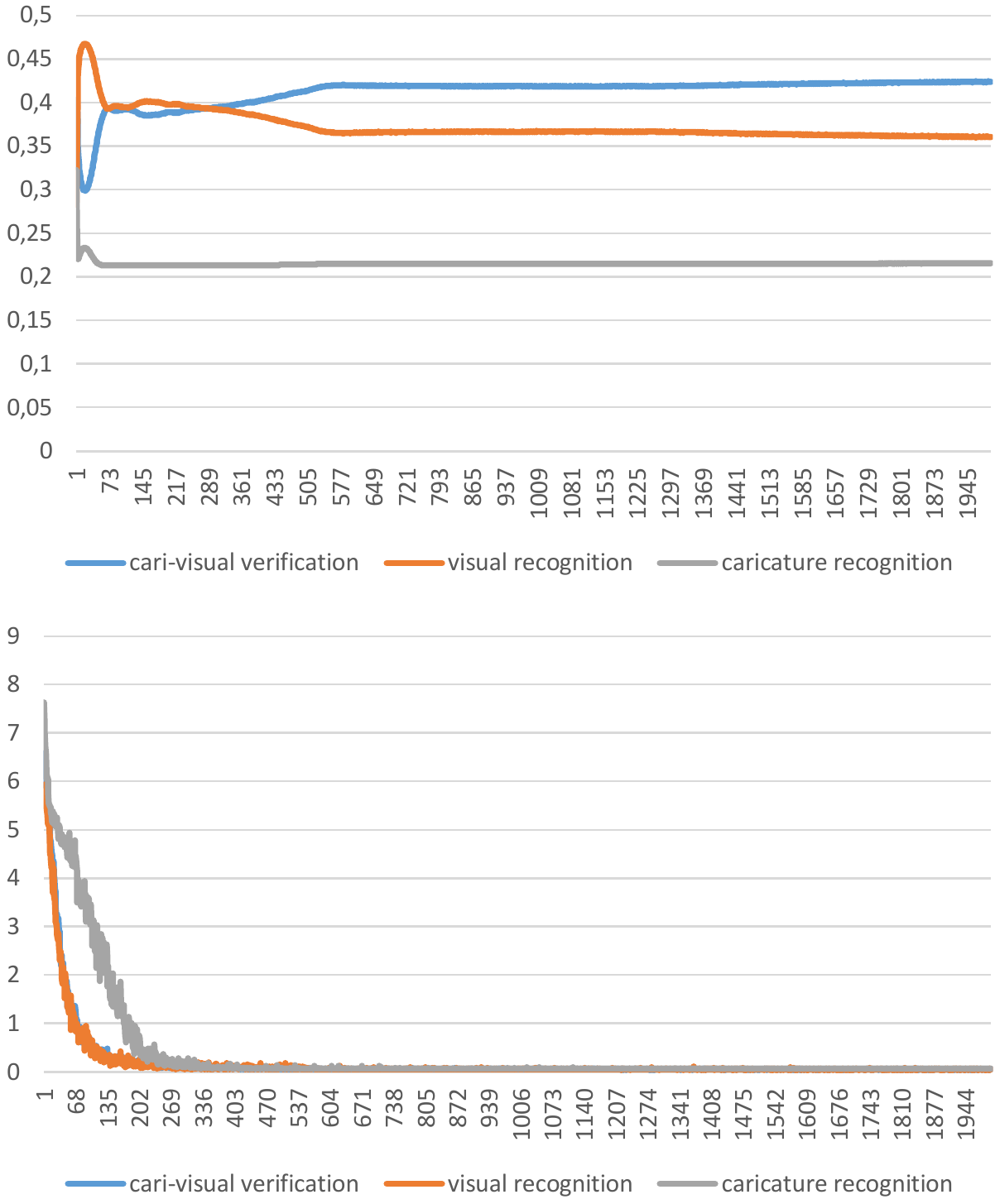}
  \subcaption{Naive Dynamic-CAVI}
  \label{fig:dynweights_real_oulu}
\end{minipage}
\begin{minipage}{.25\textwidth}
  \centering
  \includegraphics[width=\linewidth]{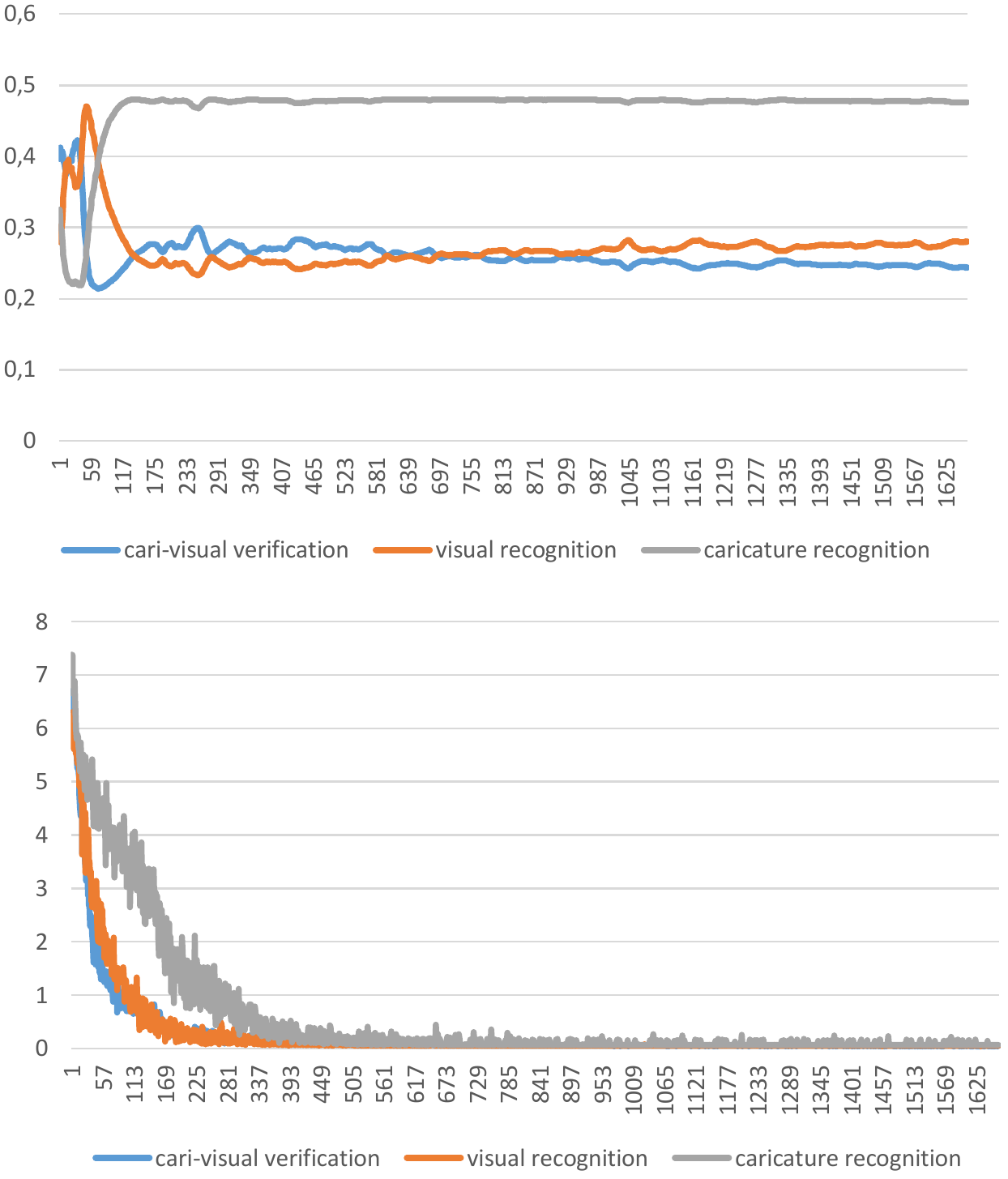}
  \subcaption{Our Dynamic-WebCari}
  \label{fig:dynweights_fake_CK+}
\end{minipage}
\begin{minipage}{.235\textwidth}
  \centering
  \includegraphics[width=\linewidth]{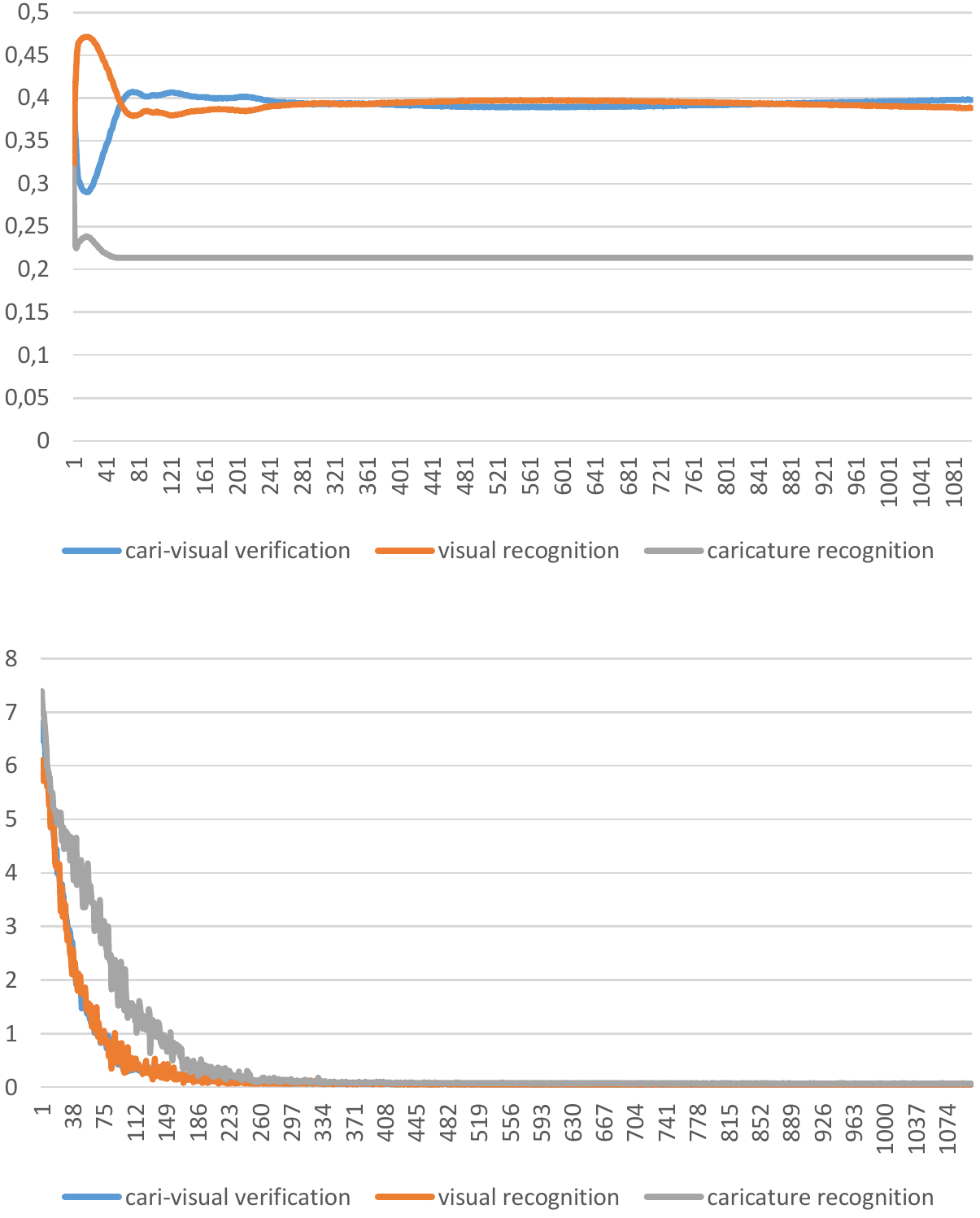}
  \subcaption{Naive Dynamic-WebCari}
  \label{fig:dynweights_fake_oulu}
\end{minipage}
\caption{Evaluation of the proposed dynamic multi-task learning and the naive dynamic multi-task learning on the datasets CaVI and WebCaricature. The upper row shows the dynamic weights for both methods and the bottom row shows the corresponding losses. The grey curves denote the caricature recognition, the orange curves denote the visual recognition and the blue curves are the caricature-visual face verification. The horizontal axis is the number of training iterations.}
\label{fig:full-dynamic-evaluation}
\end{figure*}

\subsection{Multi-task learning for caricature recognition}
In this section, we evaluate the different approaches extensively on the datasets CaVI and WebCaricature. Unlike the toy example, all the tasks are now evaluated. ~\figurename~\ref{fig:full-dynamic-evaluation} demonstrates the comparison of the proposed dynamic multi-task learning and the naive multi-task learning for the caricature recognition, visual recognition and the caricature-visual face verification on the two datasets.  As same as shown in the toy example, our method can also adapt the weights of tasks to focus on the training of the hardest task while the naive dynamic method still trains the easiest task firstly. Besides, across all the datasets and the methods, the caricature recognition with a large loss is the hardest task to train. This is reasonable since the model has been pretrained on the visual images, it is relatively easy to train the visual image related tasks rather than training the caricature recognition nearly from scratch. ~\tableautorefname~\ref{tab:multitaskevalCaVI} shows the evaluation results of the caricature-visual face verification, caricature identification and visual face identification on dataset CaVI. It shows that for all three tasks, the proposed dynamic multi-task learning method outperforms the state-of-art method CaVINet. We also evaluate the naive dynamic multi-task learning method based on our networks. We can see that for the hard task caricature identification, the performance of the naive dynamic multi-task learning (75.80\%) is inferior to our method (85.61\%) and also worse than the performance of the single-task model (78.20\%), it proves that the naive multi-task learning is incapable to well train the hard task. In addition, we also report the performance of the caricature identification on the visual identification model (V2C) and vice versa (C2V). Comparing to the single-task learning, it suggest that the multi-task learning framework can obtain a much better performance by virtue of the sharing hidden layers which has learned the common features across the different recognition modalities. Comparing to the C2V model, the V2C perform better since the visual identification model has been pretrained on the large visual image dataset while the caricature model only trained on the relative small dataset for caricatures.~\tableautorefname~\ref{tab:WebCariVerif},~\tableautorefname~\ref{tab:WebCariC2P} and~\tableautorefname~\ref{tab:WebCariP2C} demonstrate the evaluation results on the WebCaricature dataset. Since the methods proposed in \cite{HuoBMVC2018WebCaricature} are the baseline methods for demonstrating the benchmark WebCaricature, the performance of our methods boost significantly the results compared to the baseline approaches. All the evaluations are conducted by the 10-folds cross validation by following the evaluation protocol of WebCaricature. We can see that on all the face verification task, Caricature to Photo identification (C2P) and Photo to Caricature (P2C) identification tasks, our method has achieved the best performance. However, there is still much room to improve in terms of the weak performance of the validation rate (recall rate) at a low false accept rate (false positive rate).

 \begin{table*}
\caption{\label{tab:multitaskevalCaVI}Evaluation of different methods for caricature-visual face verification, caricature identification and visual face identification (accuracy\%) on dataset CaVI.}
\begin{center}
\small
\begin{tabular}{|l|c|c|c|c|c|}
\hline
Method & Verification & Visual-id & Cari-id &V2C& C2V\\
\hline\hline
CaVINet &91.06&94.50&85.09&-&-  \\
CaVINet(TW) &84.32&85.16&86.02&-&-  \\
CaVINet(w/o ortho) &86.01&93.46&80.43&-&-  \\
CaVINet(shared) &88.59&90.56&81.23&-&-  \\
CaVINet(visual) &88.58&92.16&83.36&-&-  \\\hline
{Navie Dynamic}  &93.80&97.60&75.80&61.90&62.80  \\
{Ours (Single-verif)}  &92.46&-&-&-&-  \\
{Ours (Single-visual)}  &-&98.10&-&-&41.80  \\
{Ours (Single-cari)}  &-&-&78.20&53.60&-  \\
{Ours (Dynamic MTL)}  &\textbf{94.92}&\textbf{98.35}&\textbf{85.61}&\textbf{80.04}&\textbf{64.39}  \\

\hline
\end{tabular}
\end{center}

\end{table*}

 \begin{table*}
\caption{\label{tab:WebCariVerif}Evaluation of different methods for caricature-visual face verification in terms of the validation rate (\%) on WebCaricature dataset.}
\begin{center}
\small
\begin{tabular}{|l|c|c|c|}
\hline
Method & VAL@FAR=0.1\% & VAL@FAR=1\% & AUC\\
\hline\hline
SIFT-Land-ITML & 5.08$\pm$1.82  & 18.07$\pm$4.72 & 0.841$\pm$0.018  \\
VGG-Eye-PCA &21.42$\pm$2.02&40.28$\pm$2.91&0.896$\pm$0.013  \\
VGG-Eye-ITML &18.97$\pm$3.90&41.72$\pm$5.83&0.911$\pm$0.014  \\
VGG-Box-PCA &28.42$\pm$2.04&55.53$\pm$2.76&0.946$\pm$0.009  \\
VGG-Box &34.94$\pm$5.06&57.22$\pm$6.50&0.954$\pm$0.010  \\\hline
{Navie Dynamic}  &38.39$\pm$4.58&79.69$\pm$1.3&0.961$\pm$0.004  \\
{Ours (Single-verif)}  &42.10$\pm$3.05&\textbf{84.52$\pm$0.80}&0.948$\pm$0.002  \\
{Ours (Dynamic MTL)}  &\textbf{45.82$\pm$1.65}&83.20$\pm$2.00&\textbf{0.987$\pm$0.002}  \\

\hline
\end{tabular}
\end{center}

\end{table*}

 \begin{table}
\caption{\label{tab:WebCariC2P} Evaluation of Caricature to Photo identification (C2P) on WebCaricature dataset.}
\begin{center}
\small
\begin{tabular}{|l|c|c|}
\hline
Method & Rank-1(\%) &Rank-10(\%)\\
\hline\hline
SIFT-Land-KCSR & 24.87 $\pm$ 1.50  & 61.57 $\pm$  1.37   \\
VGG-Eye-PCA & 35.07 $\pm$ 1.84  & 71.64 $\pm$  1.32   \\
VGG-Eye-KCSR & 39.76 $\pm$ 1.60  & 75.38 $\pm$  1.34   \\
VGG-Box-PCA & 49.89 $\pm$ 1.97  & 84.21 $\pm$  1.08   \\
VGG-Box-KCSR & 55.41 $\pm$ 1.41  & 87.00 $\pm$  0.92   \\\hline
{Navie Dynamic}  & 86.00 $\pm$ 1.70  & 98.21 $\pm$  1.08   \\
{Ours (Single-verif)}  & 85.55 $\pm$ 1.30  & 96.31 $\pm$  0.08   \\
{Ours (Dynamic MTL)}  & \textbf{87.30 $\pm$ 1.20}  & \textbf{99.21 $\pm$  1.07}   \\

\hline
\end{tabular}
\end{center}

\end{table}

 \begin{table}
\caption{\label{tab:WebCariP2C}Evaluation of Photo to Caricature (P2C) identification (C2P) on WebCaricature dataset.}
\begin{center}
\small
\begin{tabular}{|l|c|c|}
\hline
Method & Rank-1(\%) &Rank-10(\%)\\
\hline\hline
SIFT-Land-KCSR & 23.42 $\pm$ 1.57  & 69.95 $\pm$  2.34   \\
VGG-Eye-PCA & 36.18 $\pm$ 3.24  & 68.95 $\pm$  3.25   \\
VGG-Eye-KCSR & 40.67 $\pm$ 3.61  & 75.77 $\pm$  2.63   \\
VGG-Box-PCA & 50.59 $\pm$ 2.37  & 82.15 $\pm$  1.31   \\
VGG-Box-KCSR & 55.53 $\pm$ 2.17  & 86.86 $\pm$  1.42   \\\hline
{Navie Dynamic}  & 82.80 $\pm$ 1.60  & 97.81 $\pm$  0.88   \\
{Ours (Single-verif)}  & 81.70 $\pm$ 2.60  & 95.25 $\pm$  1.08   \\
{Ours (Dynamic MTL)}  & \textbf{84.00 $\pm$ 1.60}  & \textbf{99.01 $\pm$  1.2}   \\

\hline
\end{tabular}
\end{center}

\end{table}

\subsection{Analysis}
\begin{figure*}[t]
    \begin{center}
       \includegraphics[width=0.9\linewidth]{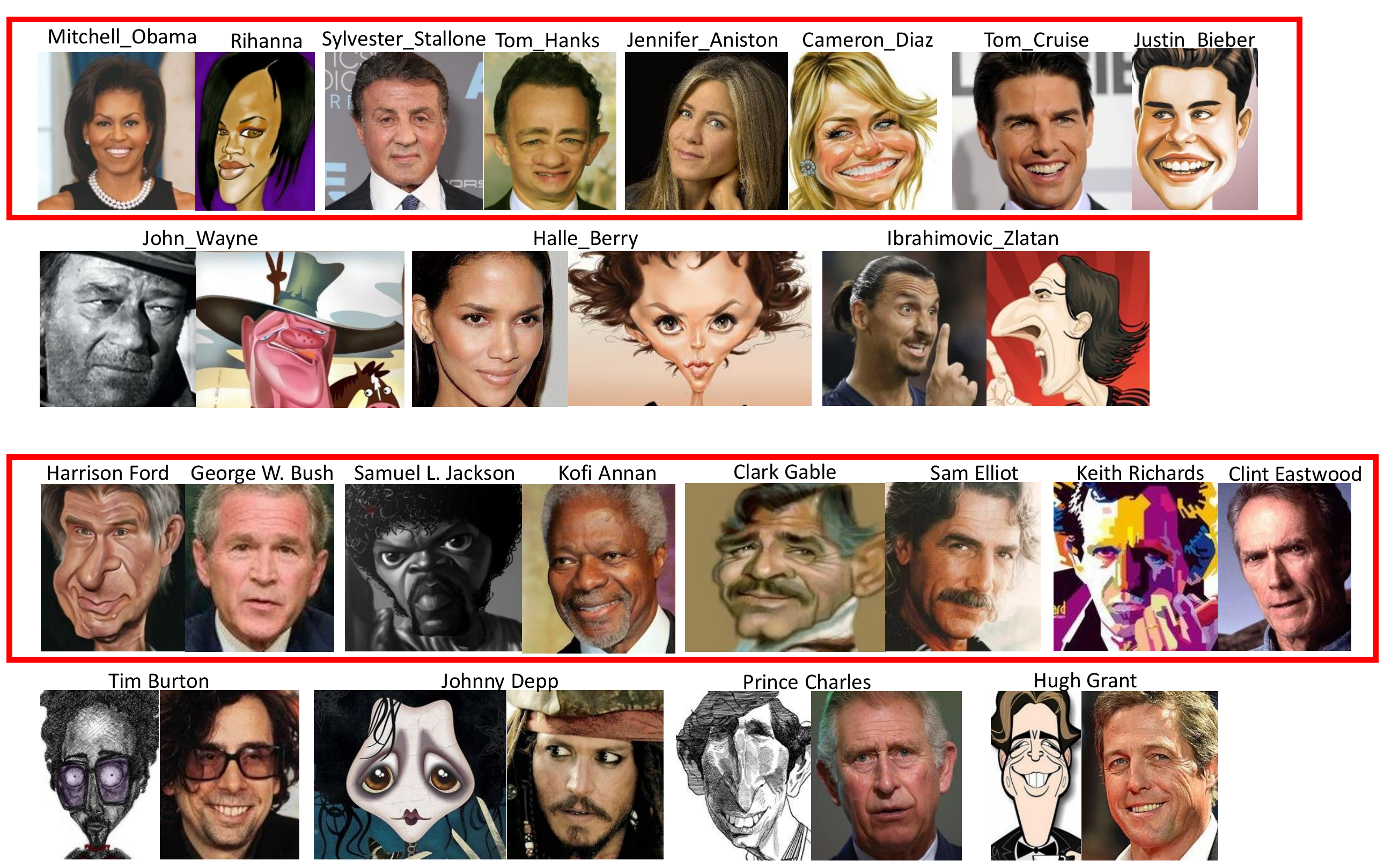}
    \end{center}
\caption{The false positive (in the red rectangles) and false negative pairs obtained by our method for caricature-visual face verification. The upper two rows are from CaVI dataset and the bottom two rows are from WebCaricature dataset.}
\label{fig:falsesample}
\end{figure*}

~\figurename~\ref{fig:falsesample} presents some false positive and false negative pairs obtained by our method from the CaVI and Web\-Caricature datasets. The false positive pairs are the pairs with the different identities mistaken recognized as the same person, while the false negative pairs are the pairs with same identities mistaken recognized as the different persons. We can see that the caricatures and the visual images in the  false positive pairs (in the red rectangles) are similar to some extent such as the si\-mi\-lar features of the pose, facial expression, hair styles, etc. However, the reason of the false negative pairs is much more diverse. The great distortion introduced by the exaggerated artistic style maybe the first reason. What is interesting is that human being can still perceive some delicate features to recognize these pairs with exaggerated distortion, it indicates that the machine could also learn to capture these features to improve the capacity for the cross-modal recognition. 

\section{Conclusion}
 In this work, we propose a multi-task learning approach with dynamic weights for the cross-modal caricature-visual face recognition, which can learn the different recognition modalities by the different tasks. Unlike existing methods, the proposed dynamic weight module without introducing the additional hyperparameters can lead the multi-task learning to train the hard task primarily instead of being stuck in the over training of the easy task. Both the theoretical analysis and the experimental results demonstrate the effectiveness of the proposed approach to learn the dynamic weights according to the importance of tasks. It also shows the superiority over the state-of-art methods for cross-modal caricature recognition. Although this dynamic multi-task learning approach is proposed for the caricature face recognition problems, it can be also easily reproduced for the other problems using deep multi-task learning based on deep CNNs by virtue of the simple structure to generate the dynamic weights of tasks. Future work will look to investigate the applicability of dynamic modules for learning the weights of task in other widely used multi-task learning frameworks such as the Mask R-CNN for image segmentation and object detection.


%
%

\bibliographystyle{unsrt}   
\bibliography{egbib_final}   

\end{document}